\documentclass[final,3p,times]{elsarticle}

\usepackage{amssymb}
\usepackage{algorithm}
\usepackage{algpseudocode}
\usepackage{amsmath}
\usepackage{graphicx}
\usepackage{subcaption}
\usepackage{booktabs}
\usepackage{multirow}
\usepackage[table]{xcolor}
\usepackage{tabularx} 
\usepackage{threeparttable}
\usepackage{caption}
\journal{a journal}

\begin{document}

\begin{frontmatter}



\title{Driving as a Diagnostic Tool: Scenario-based Cognitive Assessment in Older Drivers from Driving Video}





\author[label1]{Md Zahid Hasan}
\author[label2]{ Guillermo Basulto-Elias}
\author[label3]{Jun Ha Chang}
\author[label2]{Shauna Hallmark}
\author[label3]{Matthew Rizzo}
\author[label2]{Anuj Sharma}
\author[label4]{Soumik Sarkar}

\affiliation[label1]{organization={Department of Electrical \& Computer Engineering},
            addressline={Iowa State University}, 
            city={Ames}, 
            state={IA},
            postcode={50011},
            country={USA}}

\affiliation[label2]{organization={Institute for Transportation},
            addressline={Iowa State University}, 
            city={Ames}, 
            state={IA},
            postcode={50010},
            country={USA}}
\affiliation[label3]{organization={Mind and Brain Health Lab}, 
            addressline={University of Nebraska Medical Center}, 
            city={Omaha},
            state={NE},
            postcode={68198}, 
            country={USA}}

\affiliation[label4]{organization={Department of Mechanical Engineering},
            addressline={Iowa State University}, 
            city={Ames},
            state={IA},
            postcode={50011}, 
            country={USA}}
\begin{abstract}
We introduce scenario-based cognitive assessment in older drivers from naturalistic driving videos. In recent times, cognitive decline, including Dementia and Mild Cognitive Impairment (MCI), is often underdiagnosed due to the time-consuming and costly nature of current diagnostic methods. By analyzing real-world driving behavior captured through in-vehicle sensors, this study aims to extract ``digital fingerprints" that reveal the correlation among driving behavior, functional decline and clinical features of dementia. Moreover, modern large vision models can draw meaningful insights from everyday driving patterns across different roadway scenarios. We propose a framework that utilizes large vision models and naturalistic driving videos captured in various driving scenarios to analyze driver behavior and assess cognitive status. We leverage the strong relationship between real-world driving behavior as an observation of the current cognitive state of the drivers, where the vehicle can be utilized as a ``Diagnostic Tool". The results indicate that our approach effectively captures the link between cognitive load and behavioral patterns. This work aims to develop a scalable and non-invasive monitoring strategy for detecting cognitive decline in the aging population.

\end{abstract}

\begin{graphicalabstract}
\begin{figure}[ht]
\centering
\includegraphics[width=1.0\linewidth]{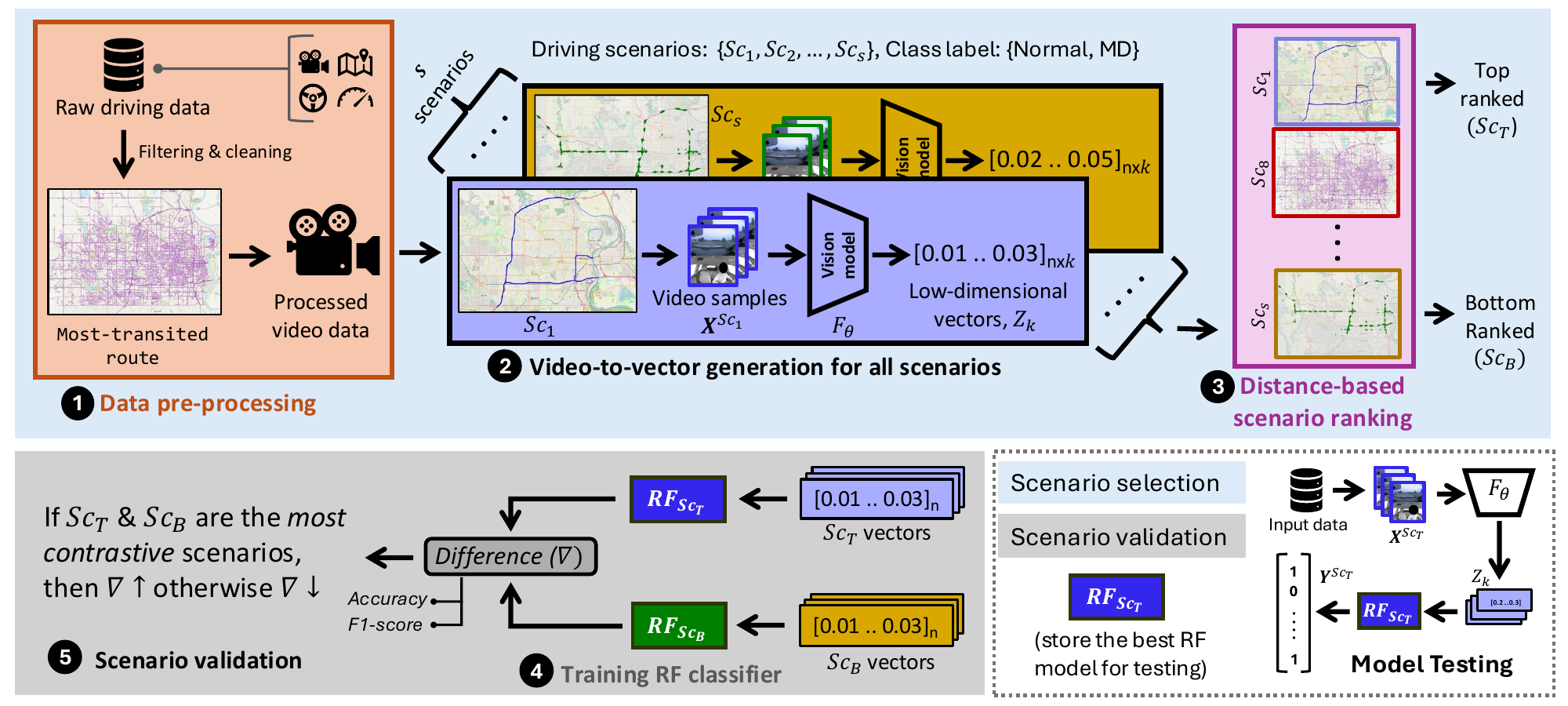}
\label{fig:graphicalabstract}
\end{figure}
\end{graphicalabstract}

\begin{highlights}
\item We introduce a novel \textbf{Scenario-based Cognitive Assessment} framework, utilizing naturalistic driving videos and large vision models to identify cognitive decline in older adults.
\item Our framework extracts discriminative visual patterns across diverse driving contexts and captures subtle behavioral cues to detect cognitive state.
\item This method demonstrates strong generalization and effectiveness in diverse driving scenarios and with unseen subjects, achieving up to 72.3\% accuracy and an F1-score of 0.80  in the most contrastive scenario.
\item We highlight naturalistic driving behavior as a scalable and effective \textbf{Diagnostic Tool} which can offer passive cognitive health assessment for early dementia detection.
\end{highlights}


\end{frontmatter}




\section{Introduction}
\label{sec:introduction}
Cognitive decline in the aging population is a critical health concern that manifests across a spectrum of stages. The pre-dementia stage, commonly referred to as Mild Cognitive Impairment (MCI), is characterized by measurable cognitive impairment with minimal or no functional changes in daily living~\cite{jamaneurol, roberts2013classification, jack2018nia}. Individuals often in their 70s may exhibit early signs of cognitive decline without severe disruptions to independence. As the condition progresses to dementia, more severe cognitive and functional impairments emerge, ultimately diminishing the ability to live independently~\cite{anderson2019state, verghese2023everyday, lin2025global}. While MCI is not exclusive to dementia, it often serves as an early indicator of the disease and highlights the need for early detection and intervention~\cite{chen2023global, ohman2021current}.

The timely identification of MCI provides an opportunity to slow disease progression, improve quality of life, and reduce the societal burden associated with advanced stages of dementia~\cite{liss2021practical, sommerlad2023social, mian2024overlooked}. Traditional methods for diagnosing cognitive decline are often time-consuming, costly, and reliant on clinical visits or complex neuroimaging techniques, which may not be practical for widespread use~\cite{liu2024detection, roberts2013classification, LASKE2015561, ALLAN2017103}. To address this, our research explores an innovative approach using real-world driving behavior as a passive diagnostic tool~\cite{bayat2021gps, rizzo2011impaired}. This study focuses on individuals exhibiting MCI due to mild dementia, as determined by clinical characteristics~\cite{jack2018nia}.

Prior research on cognitive decline has explored a variety of methods to detect early changes in dementia risk. Neuroimaging techniques, such as magnetic resonance imaging (MRI)~\cite{chouliaras2023use, jitsuishi2022searching, valizadeh2025deep}, Electroencephalography (EEG)~\cite{aljalal2024mild, aljalal2024selecting} and positron emission tomography (PET)~\cite{rallabandi2023deep, ossenkoppele2022amyloid} have been widely used to identify structural and functional brain changes associated with dementia. These methods are often paired with biomarkers, such as amyloid-beta and tau proteins~\cite{pradeepkiran2024amyloid, john2021synaptic}, blood-based biomarkers~\cite{grande2025blood, hampel2023blood} to improve diagnostic accuracy. However, the high cost and limited accessibility of these techniques hinder their widespread application~\cite{zhang2024recent}. Wearable sensors~\cite{rykov2024predicting, 10080910, whelan2022developments} have also gained traction in recent years. Studies have demonstrated the utility of devices that monitor gait patterns, heart rate variability, and sleep quality to assess functional and cognitive impairments. Machine learning models~\cite{xue2024ai, khodabandehloo2021healthxai, lin2023comprehensive} applied to sensor data have shown promise in predicting the progression of cognitive decline. However, these approaches often require participant compliance and are limited to specific contexts, such as controlled environments or clinical trials~\cite{langbaum2023recommendations, sanderson2022cognitive}.
Some studies demonstrated that everyday driving behavior, such as reduced driving distances, fewer unique destinations, and smaller driving spaces, is associated with symptomatic MCI~\cite{wang2021visual, kostyniuk2008self, davis2012road, eby2012driving}. However, limited research has investigated the impact of preclinical MCI on driving behavior~\cite{di2021using, OTT2017136}, highlighting a critical gap in understanding early-stage cognitive decline.  Recently, driving research has shifted towards naturalistic assessments using sensor devices installed in participants' personal vehicles, enabling the collection of real-world driving behavior data~\cite{bayat2021gps, camilleri2023driving, chang2023real, di2021using}.

While neuropsychological testing provides a comprehensive assessment of cognitive abilities, it often requires in-person clinical visits, participant compliance, and substantial time and resources — all of which may be challenging for individuals in the early stages of MCI~\cite{dokholyan2022challenges, massett2021facilitators}. Recent naturalistic driving research, such as the LongROAD~\cite{di2021using}, demonstrated that combining driving variables with demographics can achieve up to 88\% F1-score in predicting incident MCI or dementia. However, it still depends on aggregated trip-level statistics and lacks context-specific behavioral insights. Additionally, some recent studies~\cite{derafshi2024impact, feng2021driving, odusami2024machine, wang2021visual} have focused on scenario-dependent driving patterns analysis. For instance, cognitively impaired drivers tend to exhibit riskier behavior under adverse weather conditions~\cite{hasan2023roadway}. These studies underscore the need for analyzing scenario-specific behaviors.

Our method addresses these gaps by leveraging naturalistic video from specific driving contexts (e.g., freeway merging, stopping in a signalized intersection) and large vision models to extract nuanced, scenario-specific visual patterns. This setting offers detection of subtle behavioral cues revealed by the cognitive load of a driving scenario. Our approach highlights the importance of monitoring real-world driving behavior as a non-invasive, cost-effective, and scalable solution. Analyzing naturalistic driving data offers a promising alternative to capture subtle behavioral cues of cognitive decline in real-world settings~\cite{merickel2019real, Sayeh2025commondriving}. Studies show that behavioral patterns in everyday driving can be an alternative and effective approach to detect early-stage MCI, complementing traditional cognitive testing and potentially enabling earlier intervention~\cite{merickel2019real, Adversedriving2023Sayeh, di2021using}. By combining driving video data with advanced computer vision techniques, we aim to discover the relationship between driving behavior and cognitive assessment, providing a digital ``diagnostic tool" for early detection of cognitive decline. We propose a scenario-based framework that identifies discriminative behavioral cues and shows how scenario-specific analysis can reveal early cognitive decline more effectively than aggregate driving metrics.

The main contributions of this paper are summarized as follows: 
\begin{itemize}
    \item To address the limitations of the traditional driving metrics-based approach in detecting early cognitive decline, we proposed a novel ``Scenario-based Diagnostic Tool" that leverages naturalistic driving videos and a Large Vision model for cognitive status recognition in older adults.  
    \item To analyze the driving contexts due to cognitive decline, we extracted discriminative visual features and leveraged a scenario-ranking strategy based on inter-group separability, which directly maps to the cognitive load of the driving task under stressful scenarios. 
    \item To evaluate robustness and generalizability, we validated our framework under two sampling conditions - random sampling and driver-level separation (DLS) with cross-validation, showcasing consistent improvements in distinguishing cognitive aging groups across ``unseen" participants. 
    \item We highlighted the potential of scenario-based driving behavior monitoring as a scalable, cost-effective, and non-invasive ``Diagnostic Tool" that offers clinical interventions for cognitive health assessment. Our framework achieved a maximum accuracy of 72.32\% and a maximum F1-score of 0.80 (on the Freeway interchange scenario), outperforming a recent state-of-the-art approach on the same task. This result highlights the efficacy of scenario-based video analysis for early detection of cognitive decline.  
\end{itemize} 

The rest of this paper is organized as follows: Section~\ref{sec:Methodology} provides a comprehensive overview of the proposed method, including data processing, sampling protocols, vision model-based and scenario-based framework. Section~\ref{sec: Experimental Setup} introduces the details of our experiment setup, including the large vision model backbone, scenario ranking strategy and classification model. Section~\ref{sec:Results} describes the experimental results and their implications for scenario-based cognitive aging-group classification. Finally, Section~\ref{sec:Conclusion} concludes the paper.

\section{Methodology}
\label{sec:Methodology}

\begin{figure}[ht]
\centering
\includegraphics[width=1.0\linewidth]{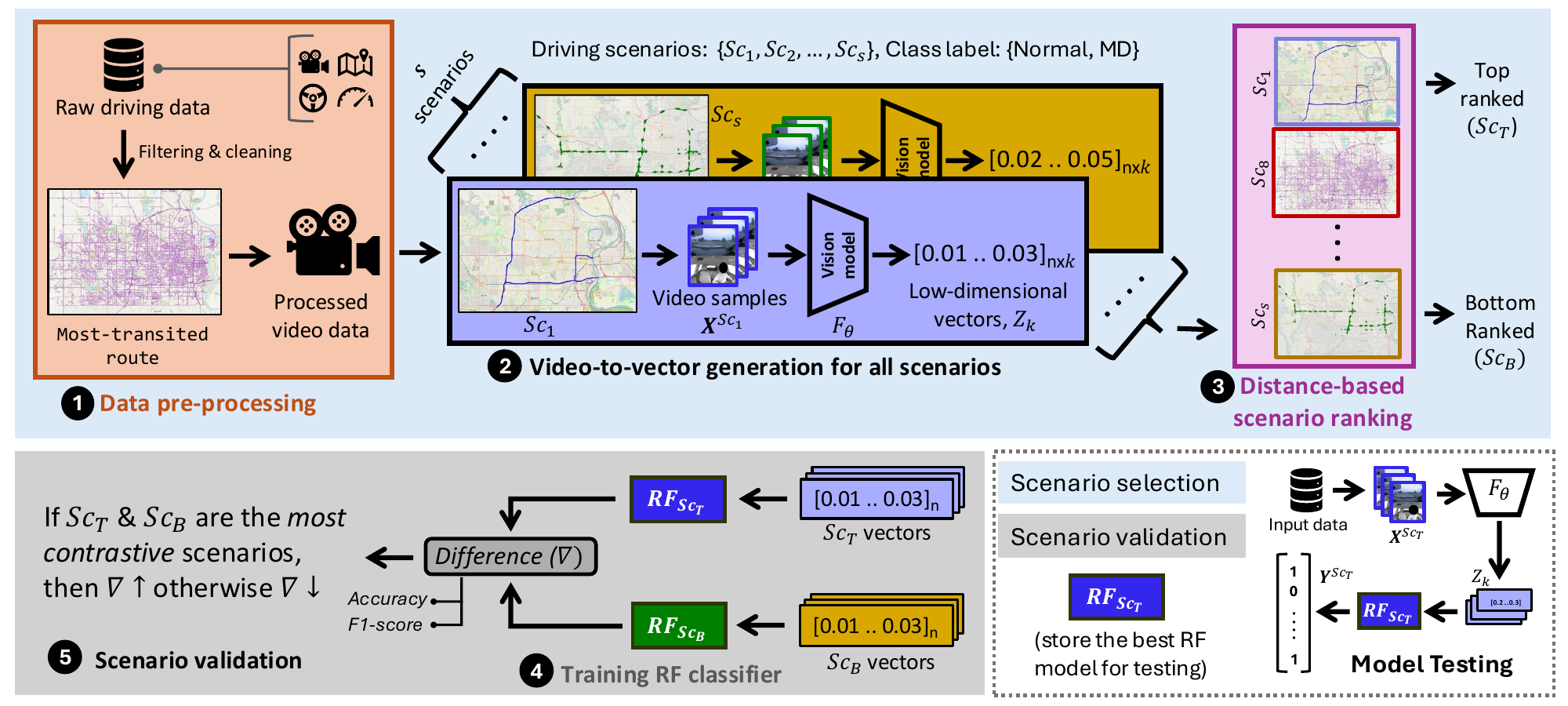}
\caption{Scenario-based cognitive assessment framework.
The pipeline consists of two phases - scenario selection and scenario validation. For testing, the fine-tuned RF model that produced the highest performance was used.}
\label{fig:DDT-framework}
\end{figure}

We proposed a scenario-based framework for cognitive status assessment from naturalistic driving videos. As illustrated in Fig.~\ref{fig:DDT-framework}, the pipeline consists of five modules. First, (1) raw driving videos were collected, filtered, and aligned to the \texttt{most-transited routes} (discussed in section~\ref{sec:video-data-process}) to ensure consistent scenario coverage across subjects. Next, (2) a pretrained large vision model encodes each video segment to vectors that were further projected to low-dimensional feature representations ($Z_k$). Then, (3) scenarios were ranked based on the Euclidean distance between the Normal and MD-aging groups in the low-dimensional space. The scenario with the largest Euclidean distance (most contrastive) and the smallest distance (least contrastive) were selected. To evaluate the ranking, (4) the most and least contrastive scenario vectors were then used to train two identical Random Forest (RF) models~\cite{breiman2001random}. Finally, (5) performance difference between these RF models was observed as they indicate the validity of the scenario-ranking and contrastive scenario selection step. A large performance gap indicates \textit{most contrastive} scenario pair. During testing, the best-performing RF model was utilized to classify the cognitive groups. In this section, we will discuss each module and model configuration in detail.

\subsection{RWRAD Dataset}
The dataset was collected in collaboration with the University of Nebraska Medical Center (UNMC) and includes comprehensive neuropsychological assessments, clinical tests and naturalistic driving data aimed at investigating cognitive impairment and its progression among older adults.

\subsubsection{Participants}
These data were derived from a cohort of community-dwelling older adults (aged 65 - 90 years) enrolled in longitudinal studies on dementia and driving behavior, conducted at the UNMC Mind and Brain Health Labs between March 2021 and January 2024. Recruitment procedures followed prior work~\cite{chang2025day, merickel2019real}, involving community outreach efforts (e.g., flyers, social media, local news, discussions at senior organizations) and pre-screening to ensure eligibility based on defined inclusion criteria (see section~\ref{appendix_a}). All participants lived independently in Nebraska or neighboring states (e.g., Iowa), held a valid driver’s license, were proficient in English and provided informed consent. Exclusion criteria were applied to minimize potential confounding effects from underlying health conditions and are detailed in section~\ref{appendix_a}. The study protocol was approved by the UNMC Institutional Review Board (IRB 0522-20-FB) and all participants provided written informed consent prior to participation. A summary overview of the participants is shown in Table~\ref{tab:participants}.

\subsubsection{Data collection}
Data collection includes recording driving video, sensor signals, actigraphy, sleep log, driving behavior and in-lab cognitive assessments conducted at UNMC. Over a two-year period, driving data were collected in three-month intervals using custom-installed “Black Box” systems in each participant’s personal vehicle. The system continuously recorded driving behavior throughout each trip over a 12-week period. It captured synchronized video (front roadway and cabin), vehicle sensor data (GPS, accelerometer, gyroscope), and on-board diagnostics data (speed, rpm, engine throttle). All data were recorded at a 1 Hz sampling rate from engine start to stop (ignition cycle). Each ignition cycle indicates a distinct ``drive" for the corresponding driver, excluding trips with zero mileage or speed $<5$mph (e.g., parking lot or driveway maneuvers) or if participants reported periods of non-driving (e.g., vacations, illness).

\begin{table}[ht]
    \centering
    \caption{Summary characteristics of the participants in the RWRAD dataset}
    \label{tab:participants}
    \resizebox{0.95\textwidth}{!}{ 
    \small
    \begin{tabular}{p{4cm} p{5cm} p{2cm} p{2cm}}  
        \toprule
        \textbf{Variable} & \textbf{Range or Counts} & \textbf{Average} & \textbf{Std. Dev.} \\
        \midrule
        Age (years) & Range: 65 - 90 & 75.4 & 6.0 \\
        \midrule
        \multirow{2}{4cm}{Sex} & Female: 75 & & \\
                               & Male: 61 & - & - \\
        \midrule
        \multirow{4}{4cm}{Race} & White: 125 & & \\
                                & Black: 9 & & \\
                                & American Indian: 1 & - & - \\
                                & Other: 1 & & \\
        \midrule
        \multirow{2}{4cm}{Income} & \$0 - \$100{,}000 & \$43{,}026.43 & \$27{,}855.35 \\
                                  & Not indicated: 86 & & \\
        \midrule
        \multirow{4}{4cm}{Employment} & Employed full/part-time: 19 & & \\
                                      & Retired: 107 & & \\
                                      & Volunteering: 9 & - &- \\
                                      & Keeping house: 1 & & \\
        \midrule
        Driving Exp. (years) & Range: 45 - 78 & 58.7 & 6.6 \\
        \midrule
        Weekly Driving (miles) & Range: 5 - 300 & 85.9 & 75.2\\
        \midrule
        COGSTAT Score & Range: 300 - 800 & 643.22 & 106.52 \\
        \midrule
        MoCA Score & Range: 15 - 30 & 24.3 & 3.4 \\
        
        \bottomrule
    \end{tabular}}
\end{table}

\subsubsection{Cognitive Status}
Cognitive status was determined using the 2018 NIA-AA research criteria~\cite{jack2018nia}. To assess cognitive status, participants completed a series of tests across five domains -- memory, language, visuospatial abilities, executive function and attention (see section~\ref{appendix_b}). Specialist clinicians in dementia and Alzheimer’s reviewed the neuropsychological evaluations for cognition and daily functioning from the National Alzheimer’s Coordinating Center (NACC) Unified Data Set (UDS) version 3~\cite{albert2011diagnosis}. They reached a consensus on cognitive staging based on participants performance across the five cognitive domains. Raw test scores were converted into standardized $Z$-scores, adjusted for age, sex, and education. These $Z$-scores were then used to define cognitive impairment. As cognitive impairment progresses, it often becomes a multi-domain process, ultimately impacting functional abilities and transitioning to the dementia stage. The in-lab data collection focused on assessing functional abilities and activities through driving behavior in various roadway scenarios. Cognitive performance was evaluated using two composite cognitive scores. 
First, the COGSTAT score was utilized as a measure of cognitive function~\cite{rizzo2009change}. It was derived from 14 neuropsychological tests evenly distributed across those five cognitive domains. For each test, an age-adjusted $Z$-score was calculated and transformed into a $T$-score (mean = 50, SD = 10) using the formula $T = 10(Z \times 10 + 50)$. The composite scores required complete data without missing tests to ensure validity. 
Second, the Montreal Cognitive Assessment (MoCA) score was used as a clinical screening tool designed to detect clinically significant cognitive change and age-related cognitive decline~\cite{nasreddine2005montreal}. Both the MoCA and the full neuropsychological battery were administered to all participants. To address challenges associated with compliance and self-reporting accuracy in cognitively impaired participants, a study partner was recruited for each participant. After two weeks, participants returned to the study site with their partner for evaluations of cognitive status and daily functioning.

Participants were classified based on the presence of both cognitive and functional decline. Those without any impairment in cognition or daily functioning were labeled as cognitively unimpaired (CU) or ``Normal-aging". Individuals who demonstrated cognitive decline but maintained functional independence were classified as having mild cognitive impairment (MCI). When cognitive decline was accompanied by functional impairment, participants were considered to have dementia. Given the small number of dementia cases ($N=9$), participants with MCI and dementia were combined into a single category termed Mild Dementia (MD) or ``MD-aging", similar to stage 4 in~\cite{jack2018nia}. Clinical evaluation of MCI is primarily a rule-out process aimed at determining whether the primary impairment is due to another type of dementia. However, the majority of people with Alzheimer's ($\sim$60\%) also have other types of dementia~\cite{dementiaRizzi}. While this study does not exclude individuals with co-morbid dementias, clinical assessments were designed to capture these data.

\subsection{Video Data Processing}
\label{sec:video-data-process}
\begin{figure}[ht]
\centering
\includegraphics[width=0.85\linewidth]{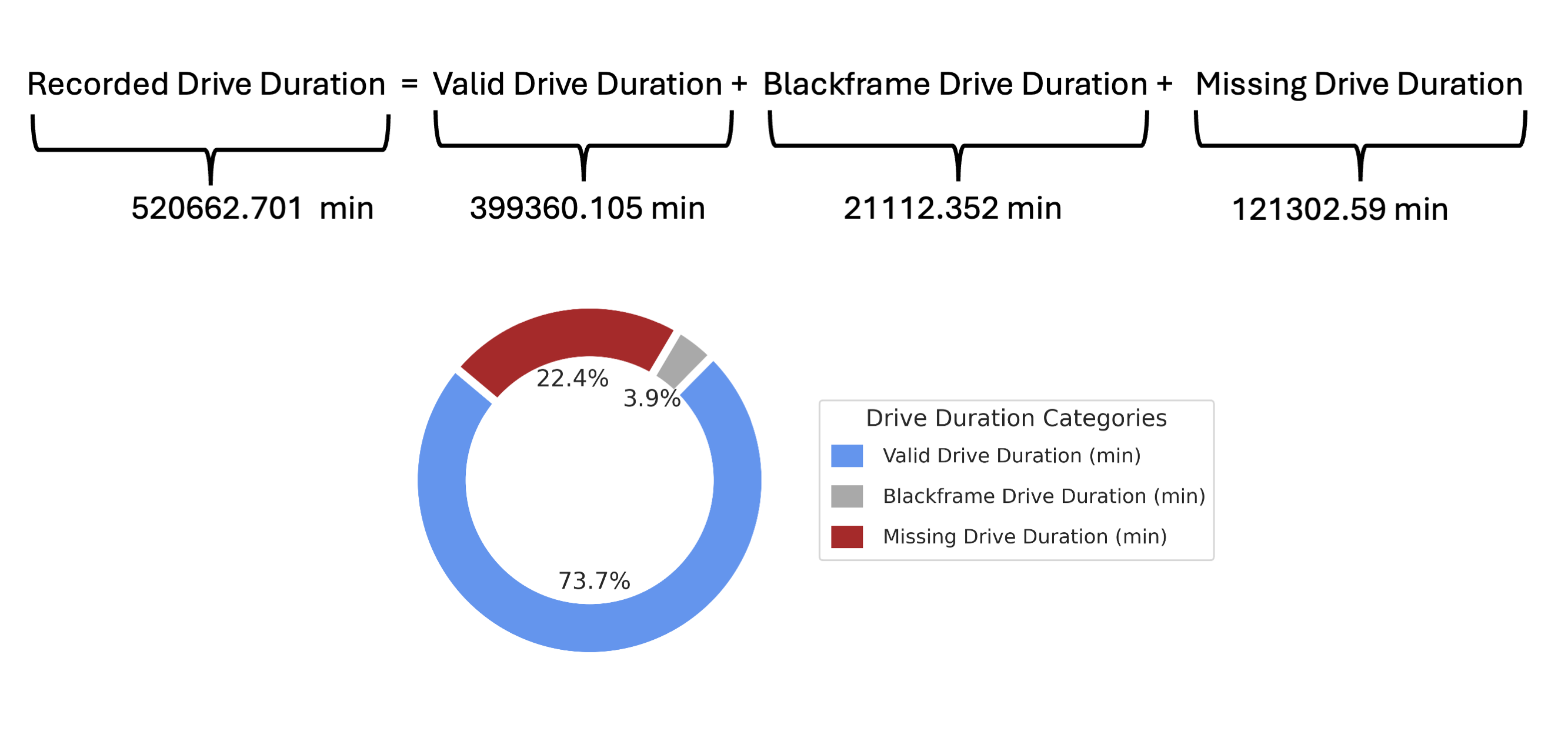}
\caption{Breakdown of the recorded driving data in the RWRAD dataset.}
\label{fig:missing-data}
\end{figure}

\begin{figure}[ht]
\centering
\includegraphics[width=1.0\linewidth]{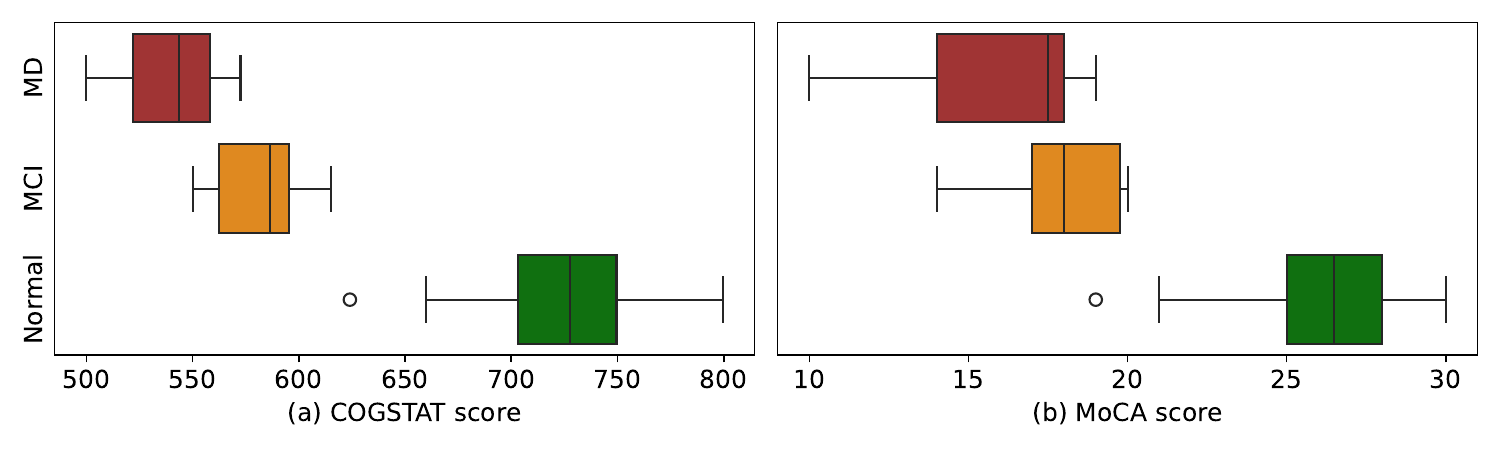}
\caption{Cognitive score distribution across all subjects ($N = 69$). (a) COGSTAT score, (b) MoCA score. We combined the MD and MCI subjects into a unified class ``MD-aging" and performed two-class classification with the ``Normal-aging".  }
\label{fig:cogstat_plot}
\end{figure}

\begin{figure}[ht]
\centering
\includegraphics[width=0.9\linewidth]{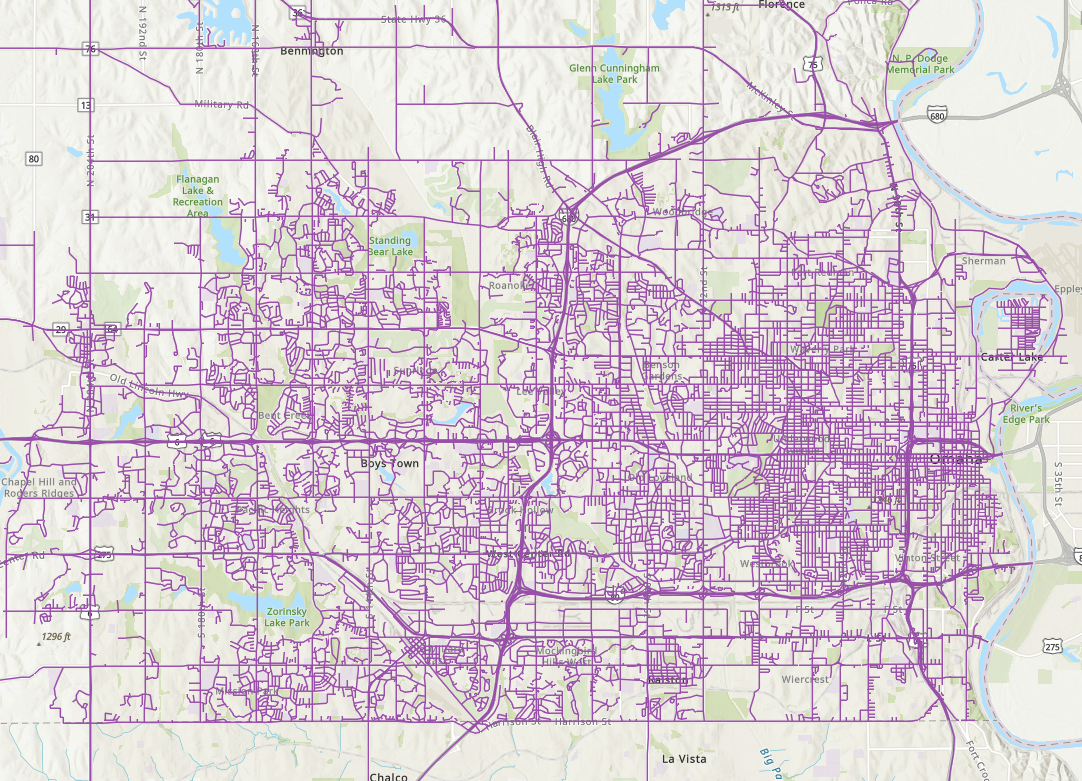}
\caption{Geospatial map of the \texttt{OMR} database (Omaha, NE). The purple lines indicate the routes where the participants drove most frequently during the RWRAD data collection period.}
\label{fig:omaha-map}
\end{figure}

\begin{figure*}[ht]
\centering
\includegraphics[width=1.0\linewidth]{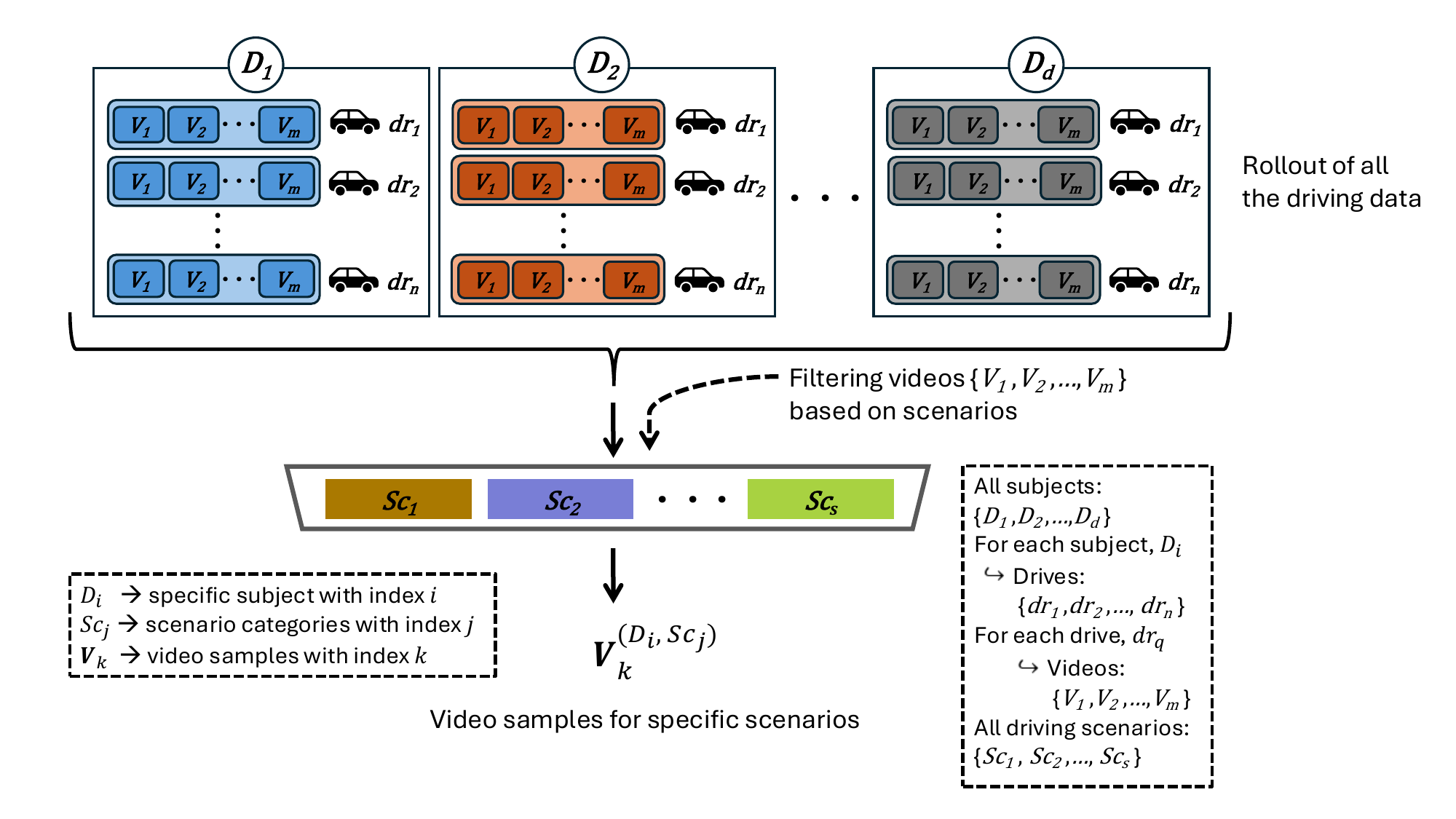}
\caption{Scenario-specific video sampling. The top row represents the rollout of all valid drives and driving videos across subjects in the \texttt{OMR} data. From each subject $D_i$, valid drives \{$dr_1, dr_2, \dots, dr_n$\} and their associated video samples \{$V_1, V_2, \dots, V_m$\} were mapped into scenario-specific subsets. These video samples were aggregated and used as input to the large vision model for driving pattern analysis.}

\label{fig:video_sample}
\end{figure*}

\begin{figure*}[h]
\centering
\includegraphics[width=1.0\linewidth]{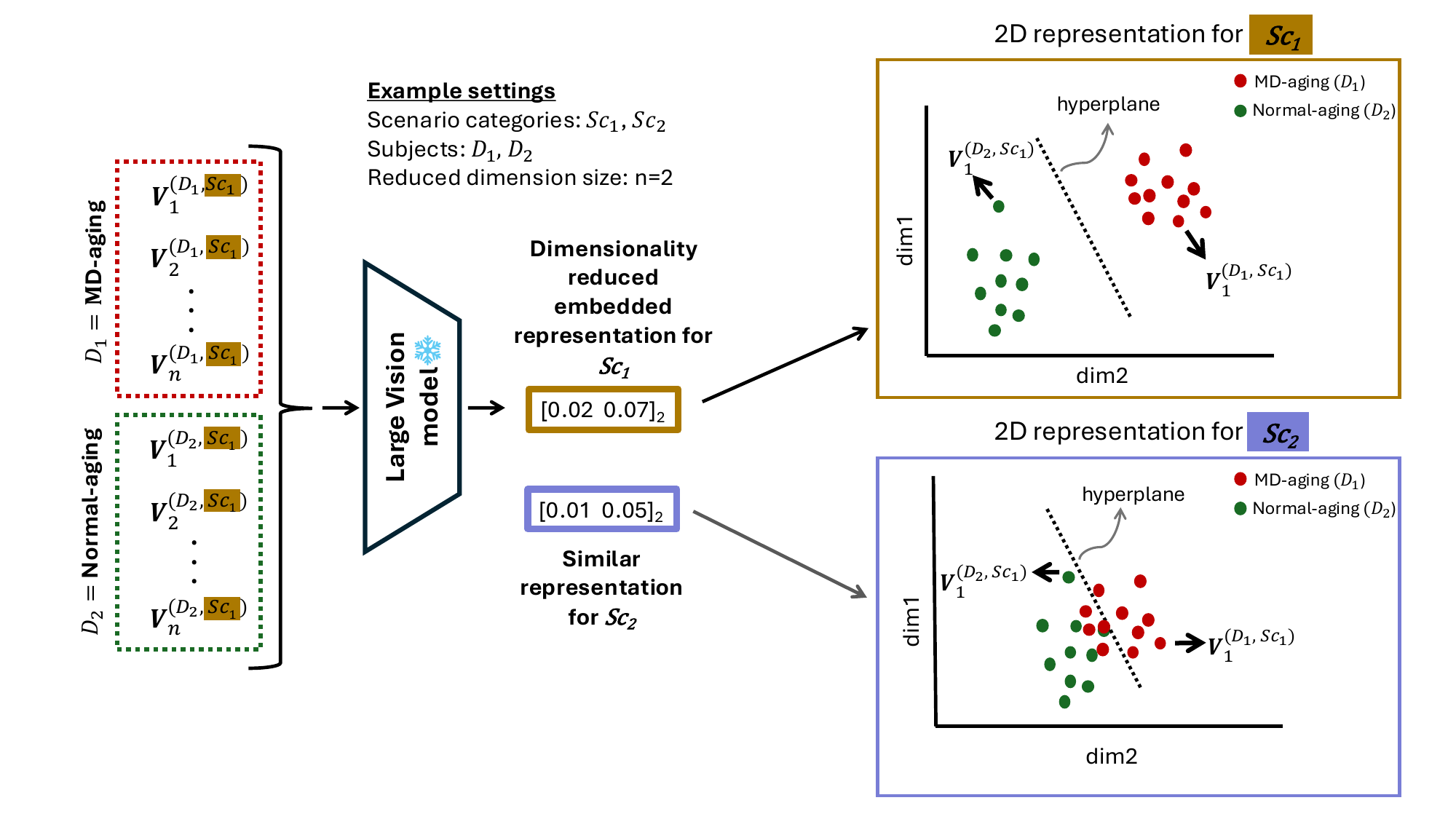}
\caption{Overview of the large vision model-based framework. Given, the input video segments $\mathcal{V} = \{V_k^{(D_i, Sc_j)}\}$ where $i=1,\dots,d;\; j=1,\dots,k;\; k=1,\dots,n$ 
from different scenarios $Sc_j \in \mathcal{S}$, a vision model $F_\theta$ mapped each $V_i$ to a vector space $Z_i = F_\theta(V_i) \in \mathbb{R}^m$. The vector space was analyzed to assess inter-group separation using a distance metric. Scenario $Sc_1$ showed significant class-wise separation than $Sc_2$, i.e., $\text{Dist}_{Sc_1} > \text{Dist}_{Sc_2}$, indicating higher discriminative potential for distinguishing the cognitive groups.
}
\label{fig:video-to-vector}
\end{figure*}

Quality driving video is essential for reliable video-based modeling since missing or corrupted data can hurt the validity of findings. Ensuring rigorous processing of the collected video clips was therefore a crucial step before conducting our analysis. While processing the collected driving video data, we encountered some missing and corrupted video clips that required filtering and formatting. Fig.~\ref{fig:missing-data} presents the breakdown of the processed driving data for the RWRAD dataset. The in-vehicle \texttt{Blackbox} system recorded a total of 520,662.7 minutes of driving data, segmented into valid drive duration (73.7\%), Blackframe drive duration (3.9\%) and Missing drive duration (22.4\%). For the video-based analysis, only valid driving data was retained. Drives with missing or corrupted data and Black-frame data were excluded to ensure quality and consistency of the input data.

Subjects with their corresponding driving videos were further filtered out for only those drives that occurred on the \texttt{Omaha Most-transited Route (OMR)}, a predefined set of road segments derived from the RWRAD dataset (shown in Fig.~\ref{fig:omaha-map}). This criterion was applied to ensure adequate video data coverage for each participant and to make sure we have diverse driving scenario exposure in the video samples. In addition, the common driving route ensures that individual drivers commute on a shared set of road segments, reducing variability due to route differences. This allows the model to focus on relevant driving patterns rather than route features. For our analysis, a total of $N=69$ subjects were filtered out, after excluding the subjects without cognitive status labels and drivers outside the Omaha region. Fig.~\ref{fig:cogstat_plot} shows the distribution of the cognitive scores across three aging groups— MD, MCI, and Normal. Among 69 subjects, 35 drivers (26 MCI and 9 MD combined) were categorized as ``MD-aging", while 34 were classified as ``Normal-aging". This balanced distribution enabled a robust comparison between normal and MD-aging groups in terms of cognitive test scores and driving behavior. The MD-aging subjects consistently show the lowest cognitive scores, while normal-aging individuals cluster at the higher end. MCI subjects span the intermediate range, overlapping with both groups. It also highlights the transitional nature of cognitive decline and the inherent challenge in classification.

To ensure consistency and relevance in our analysis, we focused on short driving segments. We selected roadway segments where the drive distance is $ \leq 0.31$ mile or $ \leq 500$ meters and defined them as  ``route segment" (discussed in section~\ref{scenario-based Roadway Selection}). These segments provide concentrated snapshots of driver behavior under controlled road conditions. This approach minimizes variability due to changes in driving context or prolonged periods of noisy observation. Additionally, the time of day and traffic conditions (peak vs. off-peak hours) were considered to account for environmental stressors that could uniquely challenge drivers with mild dementia. Next, the scenario-specific video sampling method is described in section~\ref{sub:scenario-specific-vid-sampling}.

\subsection{Scenario-specific Video Sampling}
\label{sub:scenario-specific-vid-sampling}

The overall pipeline for scenario-specific video sampling is illustrated in Fig.~\ref{fig:video_sample}. Let $\mathcal{D} = {D_1, D_2, \dots, D_d}$ be the set of subjects that were filtered out in section~\ref{sec:video-data-process}. Each subject $D_i$ performed a set of drives \{${dr_1, dr_2, \dots, dr_n}$\} where $n$ is the maximum number of drives per subject. Each drive was further segmented into short video clips \{$V_1, V_2, \dots, V_m$\}, where $m$ is the maximum number of video clips per drive. These video clips were recorded under various driving scenarios, $\mathcal{S} = {Sc_1, Sc_2, \dots, Sc_s}$. From these video clips, a subset of scenario-specific video samples $V_k^{(D_i, Sc_j)}$ was filtered for each subject-scenario pair $(D_i, Sc_j)$ where $k$ is the index of the sampled video clip, $i$ is the index of the subject and $j$ is the index of the scenario. This resulted in a quality dataset of labeled video samples that were used as inputs to the large vision model in the next stage. This sampling strategy ensures sufficient data representation across both aging groups (Normal vs. MD) and allows the proposed method to be validated on balanced and identical driving scenarios. Additionally, the hierarchical video sampling steps across subjects, drives, and scenarios ensure a structured and scalable representation across varying cognitive profiles and driving contexts.

\subsection{Large Vision Model-based Framework}
\label{large Vision Model-based Framework}
The video samples derived in section~\ref{sub:scenario-specific-vid-sampling} were fed into a pre-trained large Vision Transformer model~\cite{kolesnikov2020big} to generate vectorized representations of the video data. The objective of this framework is to encode driving videos into discriminative vector representations for downstream classification of cognitive aging groups. These representations capture prominent visual features that contrast the driving behavior of the two cognitive groups and distinguish them based on the vector's distance in feature space (illustrated in Fig.~\ref{fig:video-to-vector}). The entire scenario-based video analysis framework is shown in Algorithm~\ref{alg:scenario} which includes the following four steps:\\

\textbf{Step 1: Video-to-vector generation}\\
Each video sample $V_k^{(D_i, Sc_j)}$ is passed through a pre-trained large vision model $F_{\theta}$ to obtain a fixed-size vector space:

\begin{equation}
Z_k = F_{\theta}(V_k^{(D_i, Sc_j)}), \quad Z_k \in \mathbb{R}^m
\label{eq:video_to_vector}
\end{equation}

where $Z_k$ denotes the vector representation of each video sample $V_k^{(D_i, Sc_j)}$ in an $m$-dimensional latent space.

\textbf{Step 2: Dimensionality Reduction}\\
To interpret the structure of the vector space and the separability of the aging groups in each scenario setting, we apply PCA~\cite{WOLD198737}, a dimensionality reduction function  $g: \mathbb{R}^m \rightarrow \mathbb{R}^n$ to obtain:

\begin{equation}
\tilde{Z}_k = g(Z_k), \quad \tilde{Z}_i \in \mathbb{R}^n, 
\quad n \in \{50, 100, 200\}, \; n \ll m
\label{eq:pca}
\end{equation}

This step was performed in a way to preserve the majority of the information embedded in the high-dimensional video representation while mapping it into a low-dimensional vector space. It allows us to observe how the visual features related to the driving behavior in different scenarios for the two aging groups are clustered in a low-dimensional feature space.

\vspace{0.3em}
\textbf{Step 3: Discriminative Scenario Selection}\\
To evaluate the relevance of a scenario $Sc_j \in \mathcal{S}$ for distinguishing the two aging groups, we computed a class-wise Euclidean distance in the reduced feature space. In Fig.~\ref{fig:video-to-vector}, an example of a dimension-reduced representation ($n=2$) with the group separability in feature space is illustrated. The Euclidean distance between the class centroids was obtained as follows:

\begin{equation}
\boldsymbol{\mu}_{\text{normal}}^{(Sc_j)} =
\frac{1}{|\tilde{Z}_{\text{normal}}^{(Sc_j)}|}
\sum\limits_{\mathbf{z}\in\tilde{Z}_{\text{normal}}^{(Sc_j)}} \mathbf{z}
\label{eq:mu_normal}
\end{equation}

\begin{equation}
\boldsymbol{\mu}_{\text{MD}}^{(Sc_j)} =
\frac{1}{|\tilde{Z}_{\text{MD}}^{(Sc_j)}|}
\sum\limits_{\mathbf{z} \in \tilde{Z}_{\text{MD}}^{(Sc_j)}} \mathbf{z}
\label{eq:mu_MD}
\end{equation}

Here, $\tilde{Z}_{\text{normal}}^{(Sc_j)}$ and $\tilde{Z}_{\text{MD}}^{(Sc_j)}$ denote the sets of low-dimensional vectors under scenario $Sc_j$ for the Normal and MD-aging groups respectively. Each centroid 
$\boldsymbol{\mu}_{\text{normal}}^{(Sc_j)}, \boldsymbol{\mu}_{\text{MD}}^{(Sc_j)} \in \mathbb{R}^n$ 
represents the average position of the corresponding group in the feature space for the same scenario. The Euclidean distance between the two centroids was then obtained by:

\begin{equation}
\mathrm{Dist}(Sc_j) = \left\|\boldsymbol{\mu}_{\text{normal}}^{(Sc_j)}-\boldsymbol{\mu}_{\text{MD}}^{(Sc_j)}\right\|_2
\label{eq:Eu-distance}
\end{equation}

Then, $\mathrm{Dist}(Sc_j)$ is used to determine the most discriminative scenario as:
\begin{equation}
Sc_{\max} = \arg\max_{Sc_j \in \mathcal{S}} \; \mathrm{Dist}(Sc_j)
\label{eq:Sc_max}
\end{equation}

where $Sc_{\max}$ denotes the most discriminative scenario that maximizes the inter-group distance between the Normal and MD-aging groups centroids in the feature space.

\textbf{Step 4: Validation of Scenario Selection}\\
To validate the scenario selection, two identical Random Forest (RF)~\cite{breiman2001random} classifiers were trained and evaluated on the \emph{most} and \emph{least} discriminative scenarios obtained in Step~3. In addition to the most discriminative scenario, we calculated the least discriminative scenario as:

\begin{equation}
Sc_{\min} = \arg\min_{Sc_j \in \mathcal{S}} \; \mathrm{Dist}(Sc_j)
\label{eq:Sc_min}
\end{equation}

Let $\tilde{\mathcal{Z}}^{(Sc)}=\{\tilde{Z}_k^{(Sc)}\}$ denotes the reduced dimentional vectors (from Step~2) and $\mathcal{Y}^{(Sc)}=\{Y_k^{(Sc)}\}$ their corresponding labels for scenario $Sc \in \{Sc_{\max}, Sc_{\min}\}$. We trained two RF classifiers with identical hyperparameters on the reduced feature space for both scenarios $\{Sc_{\max}, Sc_{\min}\}$, defined as
$\hat{Y}_k^{(Sc)} = \tilde{H}^{(Sc)}(\tilde{Z}_k^{(Sc)}), \; \tilde{H}^{(Sc)} : \mathbb{R}^n \to \mathcal{Y}$. A balanced train-test split and a $K$-fold cross-validation step were used to balance the two aging groups data for training the RF classifiers. Then the trained RF models performance was assessed on classification accuracy\( \mathbf{(a)} \), precision\( \mathbf{(P)} \), recall\( \mathbf{(R)} \) and F1-score\( \mathbf{(F1)} \) computed on the test data for each scenario. The entire scenario-based workflow including the five modules from Fig.~\ref{fig:DDT-framework} is illustrated in Algorithm~\ref{alg:scenario}. Finally, we compared \(\{\mathbf{a},\,\mathbf{P},\,\mathbf{R},\,\mathbf{F1}\bigr\}_{Sc_{\max}}\) vs. \(\{\mathbf{a},\,\mathbf{P},\,\mathbf{R},\,\mathbf{F1}\bigr\}_{Sc_{\min}}\) and reported the performance difference as a validation criterion.

\begin{algorithm}[H]
\small
\caption{Scenario-Based Video Analysis}
\label{alg:scenario}
\begin{algorithmic}[1]
\State \textbf{Input:} Video samples $\mathcal{V}=\{V_k^{(D_i,Sc_j)}\}$, subjects $\mathcal{D}$, 

scenarios $\mathcal{S}$, vision model $F_\theta$, PCA $g:\mathbb{R}^m\!\to\!\mathbb{R}^n$, 

sampling method $\texttt{S}\in\{\text{Random},\text{DLS}\}$
\State \textbf{Output:} Scenario ranking $\{\mathrm{Dist}(Sc_j)\}_{Sc_j\in\mathcal{S}}$, 

trained RFs $\{\mathbf{a},\mathbf{P},\mathbf{R},\mathbf{F1}\}$

\For{each scenario $Sc_j \in \mathcal{S}$} \label{line:for-scen}
    \State $\mathcal{E}_{Sc_j}\gets\emptyset$ 
    \For{each subject $D_i \in \mathcal{D}$}
        \For{each video $V_k^{(D_i,Sc_j)}$}
            \State $Z_k \gets F_\theta\!\big(V_k^{(D_i,Sc_j)}\big)$ \Comment{Step 1}
            \State $\mathcal{E}_{Sc_j}\gets \mathcal{E}_{Sc_j}\cup\{Z_k\}$
        \EndFor
    \EndFor
    \State $\tilde{\mathcal{E}}_{Sc_j}\gets g(\mathcal{E}_{Sc_j})$ \Comment{Step 2}
    \State Divide $\tilde{\mathcal{E}}_{Sc_j}$ into two class-wise sets:
    
       $\mathbf{z_{n}}=\tilde{Z}_{\text{normal}}^{(Sc_j)}$,\quad
       $\mathbf{z_{m}}=\tilde{Z}_{\text{MD}}^{(Sc_j)}$
    \State $\boldsymbol{\mu}_{\text{normal}}^{(Sc_j)}\gets\frac{1}{|\mathbf{z_{n}}|}\sum\limits_{\mathbf{z}\in\mathbf{z_{n}}}\mathbf{z}$,\quad
    $\boldsymbol{\mu}_{\text{MD}}^{(Sc_j)}\gets\frac{1}{|\mathbf{z_{m}}|}\sum\limits_{\mathbf{z}\in\mathbf{z_{m}}}\mathbf{z}$

    \State $\mathrm{Dist}(Sc_j)\gets \left\|\boldsymbol{\mu}_{\text{normal}}^{(Sc_j)}-\boldsymbol{\mu}_{\text{MD}}^{(Sc_j)}\right\|_2$ \Comment{Step 3}
\EndFor

\State $Sc_{\max} \gets \arg\max\limits_{Sc_j\in\mathcal{S}} \mathrm{Dist}(Sc_j)$ \Comment{most discriminative scenario}
\State $Sc_{\min} \gets \arg\min\limits_{Sc_j\in\mathcal{S}} \mathrm{Dist}(Sc_j)$ \Comment{least discriminative scenario}

\For{each $Sc \in \{Sc_{\max}, Sc_{\min}\}$} \label{line:train-loop}
    \State $(\mathbf{X}^{(Sc)},\mathbf{y}^{(Sc)}) \gets (\tilde{\mathcal{E}}_{Sc}, \text{labels})$
    \If{$\texttt{S}=\text{Random}$}
        \State train/val/test split on samples (balanced)
    \ElsIf{$\texttt{S}=\text{DLS}$}
        \State Leave-$k$-drivers-out: hold out $k$ drivers as test set; 
        \State Repeat $r$ runs with different held-out drivers
    \EndIf
    \State Train RF $\tilde{H}^{(Sc)}:\mathbb{R}^n\to\mathcal{Y}$ on train split  \Comment{Step 4}
    \State Evaluate on held-out data to obtain $\{\mathbf{a},\mathbf{P},\mathbf{R},\mathbf{F1}\}_{Sc}$;
    \State Average over runs if DLS
\EndFor

\State \Return $\{\mathrm{Dist}(Sc_j)\}_{Sc_j\in\mathcal{S}}, \{\mathbf{a},\mathbf{P},\mathbf{R},\mathbf{F1}\}_{Sc_j\in\{Sc_{\max}, Sc_{\min}\}}$
\end{algorithmic}
\end{algorithm}

A consistent performance gain of $\tilde{H}^{(Sc_{\max})}$ over $\tilde{H}^{(Sc_{\min})}$ validates the proposed scenario-based approach by demonstrating that scenarios with larger inter-centroid separation yield stronger discriminability or contrast between the two cognitive aging groups. Moreover, we further evaluated the validity of the framework under different data sampling strategies. Multiple sampling strategies (random and driver-level separation) were employed to assess whether the observed RF classifier performance generalizes across subjects and settings. We reported the average performance \(\mathbf{a, P, R,}\) and \(\mathbf{F1}\) under two complementary sampling strategies:

\begin{itemize}
    \item \textbf{Random Sampling:} A total of $n$ video samples are randomly selected from the full dataset for training and testing, without enforcing subject- or scenario-level separation. This strategy evaluates the model's performance under unconstrained sampling conditions.

    \item \textbf{Driver-level Separation (DLS):} To evaluate the generalizability of the proposed framework across different individuals, we employed a leave-$k$-drivers-out cross-validation strategy~\cite{leave-out}, where $k$ drivers were excluded from the training set and reserved exclusively for testing. This procedure was repeated over $r$ independent runs, each with a different subset of left-out drivers. Finally, the results across these runs were averaged to get the final performance metrics \( \mathbf{(a, P, R, F1)} \). It ensures robustness and mitigates variance due to subject-specific biases.
    \end{itemize}

This sampling strategy evaluates both intra-subject consistency (random) and inter-subject generalization (DLS for unseen subjects) in the proposed modeling approach. 

\section{Experimental Setup}
\label{sec: Experimental Setup}
\subsection{Scenario-based Roadway Selection}
\label{scenario-based Roadway Selection}

To investigate the connection between driving behavior and cognitive load under different driving scenarios, we tested and selected two scenarios from the \texttt{OMR} data where older drivers exhibited the most discriminative driving patterns. These patterns can serve as interpretable behavioral cues for identifying cognitive decline. The selected driving scenarios are:

\begin{figure}[ht]
\centering
\begin{subfigure}[t]{0.48\linewidth}
    \centering
    \includegraphics[width=\linewidth]{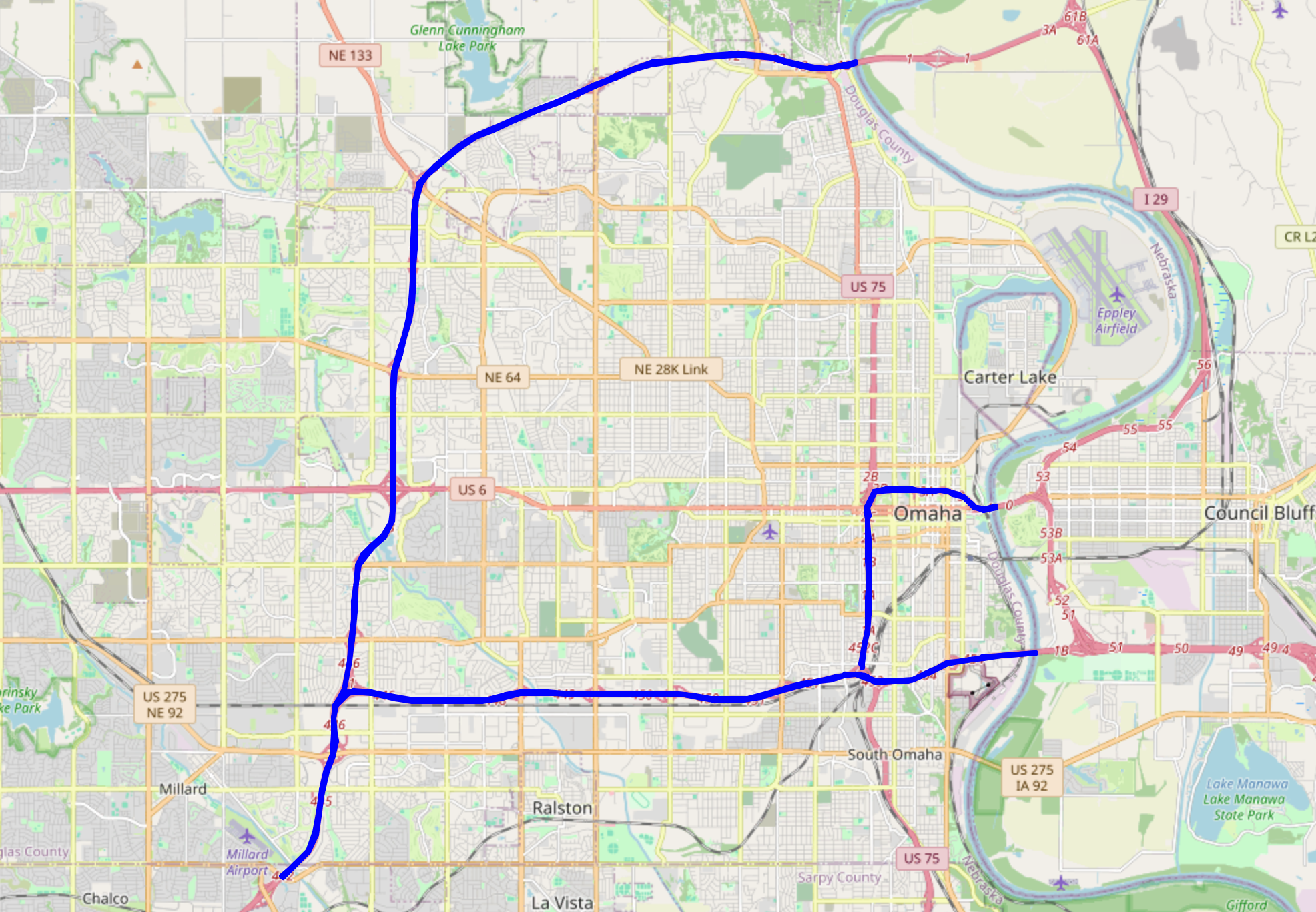}
    \caption{Interstate segments}
    \label{fig:interstate}
\end{subfigure}
\hfill
\begin{subfigure}[t]{0.48\linewidth}
    \centering
    \includegraphics[width=\linewidth]{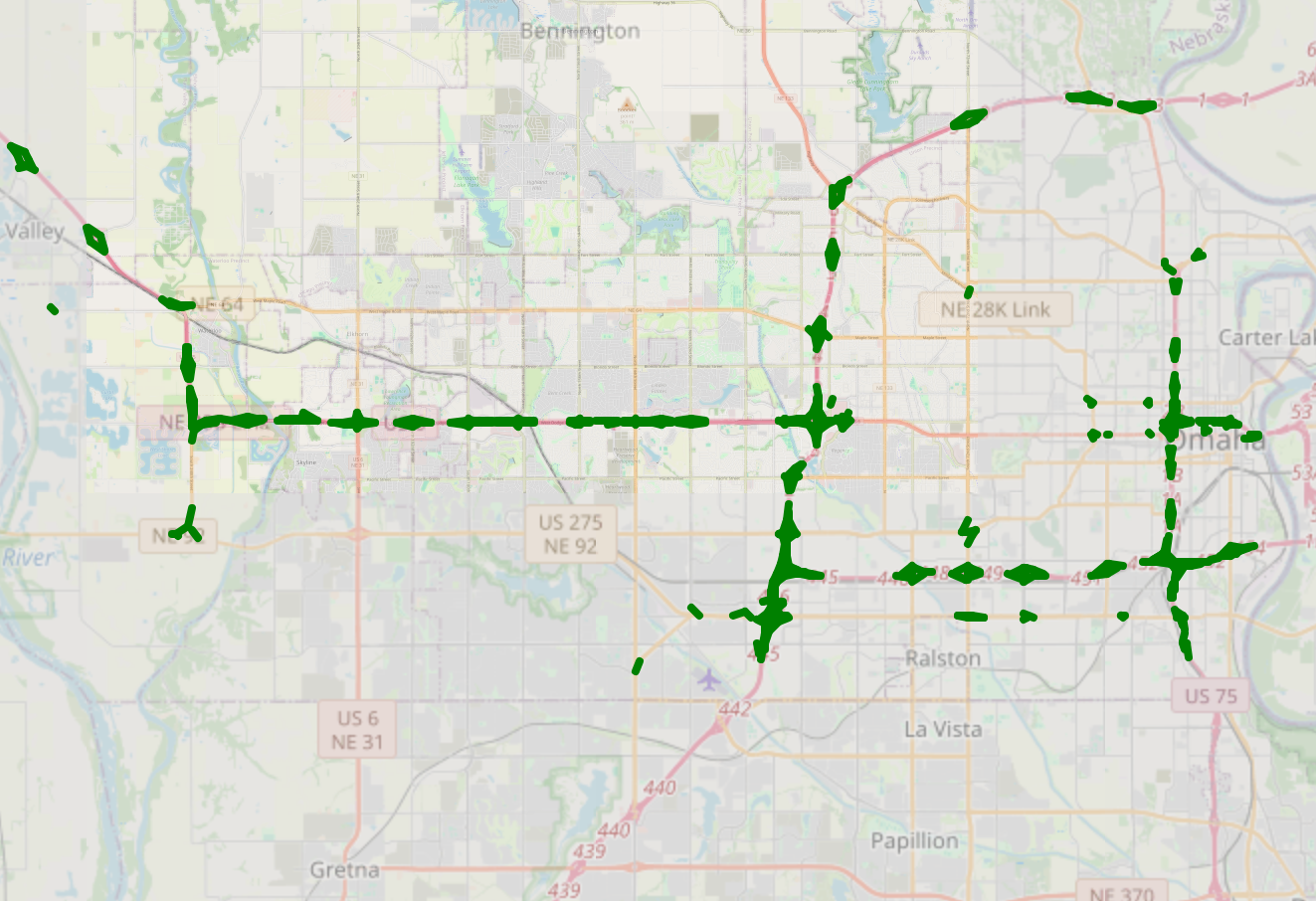}
    \caption{Freeway-interchange segments}
\end{subfigure}
\caption{GIS map for (a) Interstate and (b) Freeway-interchange scenarios. The relevant route segments were filtered out from the \texttt{OMR} database shown in Fig.~\ref{fig:omaha-map}.}
\label{fig:interstate_Fwy-int_GIS}
\end{figure}

\begin{itemize}
    \item \textbf{Freeway-interchange}: merging or diverging behavior in ramp entries or exit segments. e.g., JFK N to I-80 E Ramp, I-480 W to Dodge W Ramp.
    \item \textbf{Interstate}: high-speed, lane-keeping scenarios on long and uninterrupted roads. e.g., Omaha I-80 W, I-80 E.
\end{itemize}

From different federal functional categories (e.g., Local, Major collector, Minor collector, etc.), we selected the most contrastive scenario combination and performed scenario-specific video sampling as section~\ref{sub:scenario-specific-vid-sampling}. Fig.~\ref{fig:interstate_Fwy-int_GIS} shows the GIS map highlighting the two driving scenarios for our analysis. We considered a single drive with a length of $\leq 0.31$ miles or $\leq 500$ meters as a route segment. \texttt{OMR} database consists of 24,106 route segments. In total, we filtered out $964$ unique route segments, consisting of $632$ freeway-interchange segments (highlighted in green on Fig.~\ref{fig:interstate_Fwy-int_GIS}(b)) and $332$ interstate segments (highlighted in blue on Fig.~\ref{fig:interstate_Fwy-int_GIS}(a)). Table~\ref{tab:drive_stats_summary} summarizes the key statistics -- a balance of the selected data and driving exposure of the cognitive groups. Additionally, we selected the route segments that cover at least a minimum number of drives for both cognitive groups. To interpret the driving exposure data distribution in Table~\ref{tab:drive_stats_summary}, we defined several key variables below:

\begin{table}[htb]
\centering
\caption{Driving exposure in the~\texttt{OMR} route segments}
\label{tab:drive_stats_summary}
\resizebox{0.9\textwidth}{!}{ 
\small
\begin{tabular}{*{7}{c}}
\toprule
\textbf{Scenario} & \textbf{Normal} & \textbf{MCI} & \textbf{MD} & \textbf{Unique} & \textbf{Recurring} & \textbf{Recurring trip} \\

\textbf{categories} & \textbf{trips($N_u$)} & \textbf{trips($C_u$)} & \textbf{trips($M_u$)} & \textbf{trips($T_u$)} & \textbf{trips($\overline{T}_u$)} & \textbf{factor($T_{rep}$)} \\
\midrule
Fwy-int. & 8136 & 10134 & 919 & 19189 & 68048 & 3.55 \\
\midrule
Interstate & 7123 & 7894  & 780 & 15797 & 69620 & 4.41 \\
\bottomrule
\end{tabular}}
\caption*{\footnotesize
$N_u$: unique trips by Normal group; $C_u$: unique trips by MCI group; $M_u$: unique trips by MD group; $T_u$: total unique trips; $\overline{T}_u$: total non-unique trips;
Fwy-int.: Freeway interchange.}
\end{table}

\begin{itemize}
  \item $N_u$: Number of unique trips by the Normal-aging subjects
  \item $C_u$: Number of unique trips by the subjects with MCI
  \item $M_u$: Number of unique trips by the subjects with MD
  \item $T_u$: Total unique trips across all subject groups\\
  $T_u = N_u + C_u + M_u$ \\ It does not include repeated trips which means multiple driving sessions recorded on the same route segment by the same individuals
  \item $\overline{T}_u$: Total number of trips including repeated ones (non-unique). \\ It includes repeated trips which means multiple driving sessions recorded on the same drive segment by the same individual.
  \item $T_{rep}$: The average number of times each unique trip was repeated per subject on a route segment, defined as $T_{rep} = \frac{\overline{T}_u}{T_u}$. It gives an approximation of how often participants repeated drives on a certain route segment. For example, each unique interstate trip was repeated 4.41 times, compared to 3.55 times for freeway interchange trips. Therefore, participants tended to repeat trips on interstate route segments more frequently than on the freeway interchanges.

\end{itemize}


We leveraged ``unique drives" vs. ``repeated drives" per subject across all route segments to indicate statistical reliability and diversity in driving exposure by the two cognitive groups. The recurring trips per subject suggest that each participant had substantial driving trips with the selected roadway scenario which offers a significant amount of representative video data for our analysis.

\begin{figure}[ht]
\centering
\includegraphics[width=0.9\linewidth]{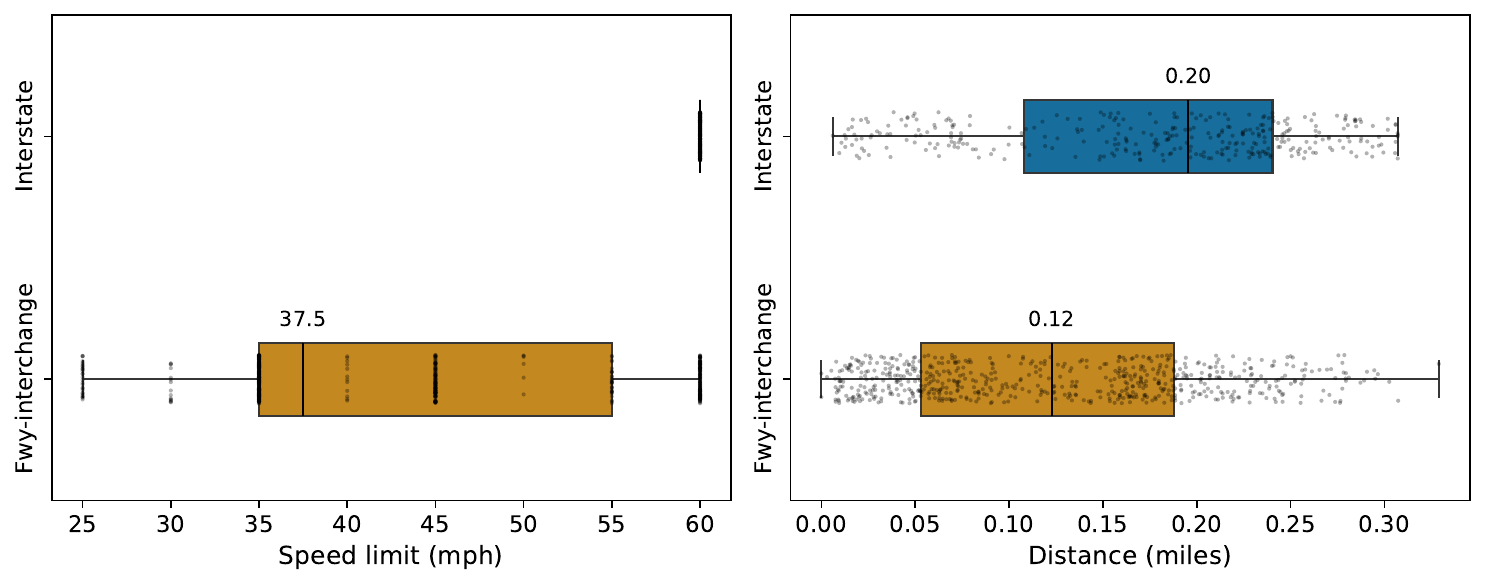}
\caption{Boxplot distributions of the speed limit and trip distance for Interstate and Freeway-interchange roadways. The middle line shows the median speed limit and median route distances.}

\label{fig:speed-dist-Int_vs_Fwy}
\end{figure}

Fig.~\ref{fig:speed-dist-Int_vs_Fwy} shows a boxplot distribution of speed limits and the trip distance for both scenarios. As we observed, the freeway-interchange segments tend to have lower speed limits and shorter distances, which work as a transitional driving scenario, involving challenging driving maneuvers and fast decision-making. In contrast, the interstate segments have higher speed limits and longer travel distances, which indicate an uninterrupted highway driving scenario with relaxed driving behavior.

\begin{itemize}
    \item Median speed limits: \\$\text{Fwy-interchange scenario}$, $S_{\text{fwy}} = 37.5$ mph \\$\text{Interstate scenario}$, $S_{\text{int}} = 60$ mph
    \item Median distances: \\$\text{Fwy-interchange scenario}$, $D_{\text{fwy}} = 0.12$ miles\\ $\text{Interstate scenario}$, $D_{\text{int}} = 0.20$ miles
\end{itemize}

These characteristics offer a strong contrast between the two scenarios. In addition, they indicate distinct driving behaviors and cognitive load associated with each scenario. As suggested in~\cite{Danda2025}, freeway merging requires substantially higher cognitive demand than straight driving due to rapid integration of sensory processing, decision-making and cognitive responses to adapt to fast-moving traffic and sudden speed changes.

\subsection{Implementation of Video-based Model}

We conducted the experiments across two representative scenarios with both transitional and straight driving maneuvers. The aim was to investigate whether the large vision model could capture the subtle visual features to effectively distinguish driving behavior relevant for cognitive assessment. The experiments were conducted under the following sampling protocols (step 4 in section~\ref{large Vision Model-based Framework}):

\begin{itemize}
    \item \textbf{Random Sampling:} Video segments were randomly split into train and test sets without enforcing any driver-level separation.
    
    \item \textbf{Driver-level Separation (DLS):} A leave-$k$-drivers-out cross-validation strategy is employed, where $k=5$ drivers were excluded during training and used exclusively for testing. This protocol was repeated over $r=3$ independent runs with different left-out driver groups and the final metrics were determined by averaging those runs.
\end{itemize}

In our experiments, the parameters were set as: (1) Input image size (960$\times$752$\times$3), (2) Vision model backbone: ResNet-v2~\cite{he2016identity} (3) Video-to-vector dimension (6144$\times$1), (4) Frame rate $\{1, 10\}$ Hz, (5) Low-dimension size, $\mathbf{n} \in \{50, 100, 200\}$, and (6) Distance metric: Euclidean distance. This setting provided a comprehensive understanding of how video-derived feature spaces behave under varying sampling conditions and data splits.

\section{Results}
\label{sec:Results}
\subsection{Model Performance on Different Scenarios}
\definecolor{lightgray}{gray}{0.90}
\begin{table}[ht]
\begin{center}
\caption{Model performance on the \texttt{OMR} subset for two contrastive scenarios}
\label{tab:video_model_results}
\resizebox{0.9\textwidth}{!}{
\begin{tabular}{p{2cm} p{3cm} p{3cm} p{3cm} p{2cm} p{2cm}}
\toprule
\textbf{Scenario} & \textbf{Accuracy} & \textbf{Precision} & \textbf{Recall} & \textbf{F1 score} & \textbf{Sampling} \\
\textbf{categories} & \( \mathbf{(a)} \) & \( \mathbf{(P)} \) &
\( \mathbf{(R)} \) & \( \mathbf{(F1)} \) & \textbf{method} \\
\midrule
Fwy-int. & 71.03\% & 0.7597 & 0.7728 & 0.7662 & random  \\
Interstate     & 55.12\% & 0.5404 & 0.5333 & 0.5368 & random \\
\midrule
\rowcolor{lightgray}

\textbf{\( \Delta \)} & 15.91\% & 0.1293 & 0.1395 & 0.1344 & random\\
\bottomrule
Fwy-int. & \textbf{69.81 ± 2.51\%} & 0.7378 ± 0.0371 & 0.7624 ± 0.0671 & 0.7501 & DLS \\
Interstate     & 52.17 ± 1.93\% & 0.5920 ± 0.0636 & 0.5997 ± 0.0771 & 0.5958& DLS \\
\midrule
\rowcolor{lightgray}
\textbf{\( \Delta_{\text{ds}} \)} & \textbf{17.64\% $\uparrow$} & 0.1458 $\uparrow$ & 0.1627 $\uparrow$ & 0.1547 $\uparrow$ & DLS\\
\bottomrule
\end{tabular}}
\caption*{\footnotesize \textbf{\( \Delta \)}: difference w/o DLS, \textbf{\( \Delta_{\text{ds}} \)}: difference w/ DLS, Fwy-int.: Freeway interchange, DLS: Driver-level Separation}
\end{center}
\end{table}

As discussed in section~\ref{large Vision Model-based Framework}, we trained two RF classifier models to classify the aging-groups (Normal-aging vs. MD-aging) across the Freeway interchange and Interstate scenarios of the \texttt{OMR} data. Table~\ref{tab:video_model_results} shows the performance of the trained RF models. The results are reported under both random sampling and driver-level separation (DLS) protocols, with the grey rows showing the performance differences (\( \Delta \) and \( \Delta_{\text{ds}} \)).

As observed, the Freeway interchange scenario consistently outperforms the Interstate scenario across all metrics \( \mathbf{(a, P, R, F1)} \), achieving 71.03\% accuracy and an \( \mathbf{F1} \) score of 0.7662 under random sampling. The performance gaps, highlighted in the grey rows (\( \Delta \) and \( \Delta_{\text{ds}} \)), confirm that certain roadway scenarios contain more discriminative vision features for distinguishing Normal-aging and MD-aging drivers. Furthermore, including the driver-level separation (DLS) yields an increase in performance difference, which can be observed on all metrics in the \( \Delta_{\text{ds}} \) row. It surpassed the performance difference achieved without the DLS case; for example, the accuracy difference went up from \( \Delta\) = 15.91\% to \( \Delta_{\text{ds}} \) = 17.64\%, indicating a 1.73\% increment in performance difference. Therefore, the Freeway interchange is a more behaviorally informative and cognitively sensitive driving scenario, where complex maneuvers, short distances and lower speed variability have more nuanced driver responses.

\subsection{Percentage of Misclassification}

In this section, we assessed the feasibility of our Vision model-based approach by calculating the percentage of misclassification for all $N = 69$ subjects (including normal and MD). For each subject \(k\), we denote \(n_{\text{test}}^{(k)}\) represent the number of times they appeared in the test set across 10 resamples, and \(n_{\text{error}}^{(k)}\) denote the number of times they were misclassified by the RF classifier. The subject-specific misclassification rate is then calculated as:
\[
\text{\% of missclassification}^{(k)} = 100 \times \left(\frac{n_{\text{error}}^{(k)}}{n_{\text{test}}^{(k)}}\right)
\]

\begin{figure}[ht]
\centering
\begin{subfigure}[t]{0.48\linewidth}
    \centering
    \includegraphics[width=0.9\linewidth]{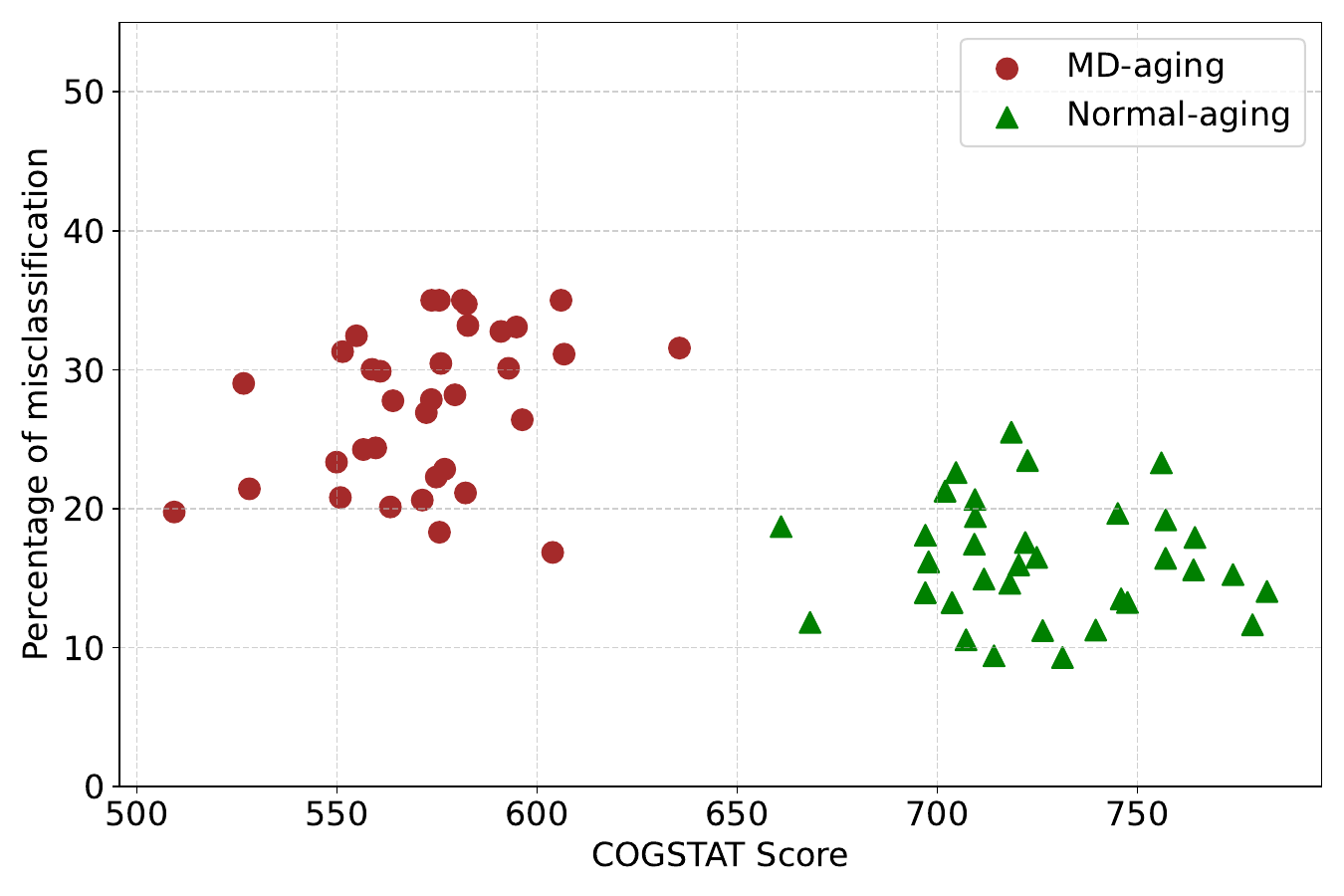}
    \label{fig:missCOGSTAT}
\end{subfigure}
\hfill
\begin{subfigure}[t]{0.48\linewidth}
    \centering
    \includegraphics[width=0.9\linewidth]{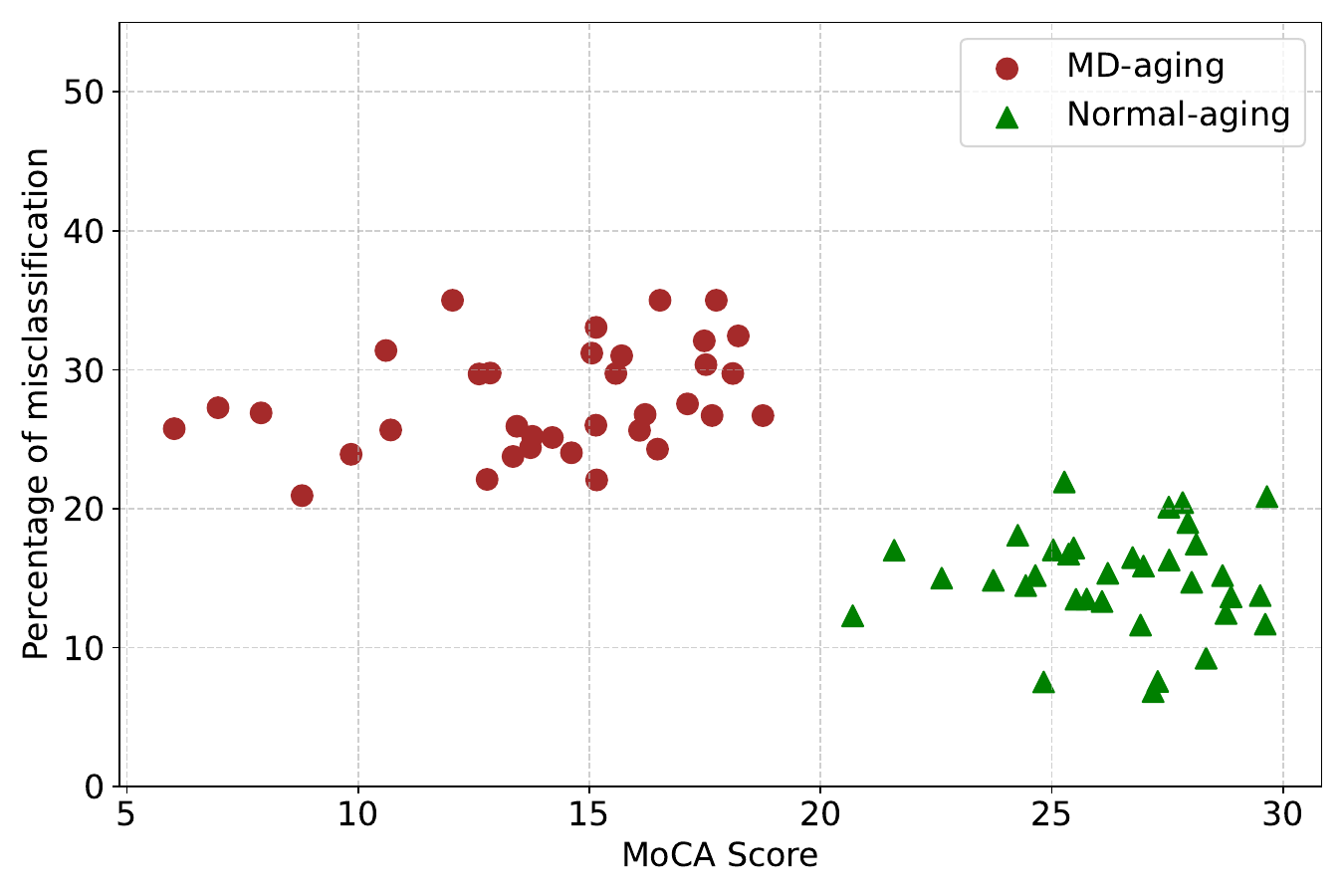}
\end{subfigure}
\caption{Percentage of misclassification vs. - (a) COGSTAT score and (b) MoCA score. The relevant route segments were filtered out from the \texttt{OMR} database shown in Fig.~\ref{fig:omaha-map}.}
\label{fig:percentage_missclassed}
\end{figure}

Fig.~\ref{fig:percentage_missclassed} displays the relationship between misclassification percentage and cognitive scores. An overall trend observed was that subjects labeled as MD-aging tend to exhibit higher misclassification percentages compared to the Normal-aging drivers. This is evident in the middle cognitive score ranges (COGSTAT: 580--650; MoCA: 12--20), where the misclassification is higher. It suggests that the middle zones correspond to a transition region where the classifier shows more uncertainty, due to the overlapping feature space between early-stage impairment and Normal-aging. On the other hand, subjects with low (COGSTAT $< 550$ and MoCA $<8$) and high (COGSTAT $> 700$ and MoCA $> 20$) cognitive scores consistently yield lower misclassification rates. This behavior indicates that the classifier learned prominent feature representations in these regions and it is able to classify the aging groups more confidently. Additionally, it reinforces the validity of using COGSTAT and MoCA as effective biomarkers in well-separated diagnostic states. Overall, these trends are consistent with the model’s performance as reported in Table~\ref{tab:video_model_results}.

\subsection{Comparison with Benchmark Approaches}

\begin{table}[ht]
\centering
\caption{Comparison with a recent benchmark tested on the RWRAD dataset}
\label{tab:comparison}
\resizebox{0.9\textwidth}{!}{ 
\begin{tabular}{p{4cm} p{4cm} p{2cm} p{2cm} p{2cm} p{2cm}}
\toprule
\textbf{Method} & \textbf{Feature(s)} &\(\mathbf{a}\) &\(\mathbf{P}\)& \(\mathbf{R}\)& \(\mathbf{F1}\) \\
\midrule
\multirow{6}{*}{Joshi et al.~\cite{joshi2025identifying}}
 & Demo & 61.86 & 70.77 & 63.89 & 67.15 \\
 & Driving      & 55.93 & 64.71 & 61.11 & 62.86 \\
 & Sleep        & 62.71 & 71.21 & 65.28 & 68.12 \\
 & Demo + Driving    & 66.95 & 71.43 & 76.39 & 73.83 \\
 & Driving + Sleep   & 63.56 & 71.01 & 68.06 & 69.50 \\
 & All  & 68.64 & 73.97 & 75.00 & 74.48 \\
 & Prediction$^*$ & 70.48 & 71.88 & 77.97 & 74.80\\
\midrule
\multirow{2}{*}{Ours} 
 & RF-Fwy-int.    & 71.03 & 75.97 & 77.28 & 76.62 \\
 & +DLS\dag   & \textbf{72.32} & \textbf{77.49} & \textbf{82.95} & \textbf{80.13} \\
  & \textcolor{blue}{\( \Delta \)} & \textcolor{blue}{+3.68} & \textcolor{blue}{+3.52} & \textcolor{blue}{+7.95} & \textcolor{blue}{+5.65} \\

\bottomrule
\end{tabular}}
\caption*{\footnotesize Demo: demographics, Driving: driving performance, Sleep: sleep log, All: five distinct features including demographics, driving performance, activity space and physiological states, $^*$cognitive status prediction for next year using previous year data, RF-Fwy-int.: RF classifier trained on \texttt{OMR} Freeway interchange routes, DLS: Driver-level Separation, \dag used max value, \textcolor{blue}{\( \Delta \)}: gain over the best performing model (All)}
\end{table}

To benchmark our framework, we compared the model performance with a recent approach proposed by Joshi et al.~\cite{joshi2025identifying} shown in Table~\ref{tab:comparison}, which includes demographics, driving features and sleep log for baseline cognitive status classification using XGBoost~\cite{hou2020predicting}. Their best-performing model achieved 68.64\% classification accuracy \( \mathbf{(a)}\) and 74.48\% \( \mathbf{F1} \) score on the RWRAD dataset. In contrast, our framework outperformed these results on the Freeway interchange scenario, achieving 71.03\% accuracy \(\mathbf{(a)}\) and 76.62\% \( \mathbf{F1} \) score under random sampling. Additionally, for the same scenario, the maximum accuracy \(\mathbf{(a)}\) and \( \mathbf{F1} \) score went up to 72.32\% and 80.13\%, respectively, with the driver-level separation (DLS) applied. These results demonstrate that scenario-based naturalistic driving video can serve as a scalable and non-invasive biomarker for cognitive aging.

\begin{table}[t]
\centering
\caption{Comparison with the recent clinical ML and DL methods}
\label{tab:clinical_comparison}
\resizebox{0.90\textwidth}{!}{
\begin{tabular}{p{4.5cm} p{3.5cm} p{3.5cm} p{2cm} p{2cm}}
\toprule
\textbf{Method} & \textbf{Modality} & \textbf{Dataset} & \(\mathbf{a}\) & \(\mathbf{F1}\) \\
\midrule
JMMLRC~\cite{8932589} 
& MRI, SNPs & ADNI-1~\cite{hinrichs2009spatially} & 58.40 & 68.30 \\

\midrule
MRN~\cite{zhang2023multi} 
& sMRI & ADNI-1,2~\cite{hinrichs2009spatially} & 63.23 & 60.23 \\
\midrule
LDA-ELM~\cite{lin2021multiclass} 
& MRI, PET & ADNI~\cite{hinrichs2009spatially} & 66.70 & 64.90 \\
\midrule
SSMFS~\cite{zhang2024self} 
& MRI, PET & ADNI-1~\cite{hinrichs2009spatially} & 67.80 & 60.42 \\
\midrule
XADLiME~\cite{mulyadi2023estimating} 
& sMRI & ADNI-1,2~\cite{hinrichs2009spatially} & 69.05 & 77.47 \\
\midrule
LA-GMF~\cite{xu2024interpretable} 
& sMRI & ADNI~\cite{hinrichs2009spatially} & 70.33 & 67.22 \\
\midrule

RF-Fwy-int. (ours) & Video & RWRAD & 71.03 & 76.62 \\
+ DLS\dag (ours) & Video & RWRAD & \textbf{72.32} & \textbf{80.13} \\
 & & \textcolor{blue}{\( \Delta \)} & \textcolor{blue}{+1.99} & \textcolor{blue}{+12.91} \\
\bottomrule
\end{tabular}
}
\caption*{\footnotesize ADNI: Alzheimer’s Disease Neuroimaging Initiative, SNP: single nucleotide polymorphisms, sMRI: structural MRI, PET: Positron Emission Tomography, RF-Fwy-int.: RF classifier trained on \texttt{OMR} Freeway interchange routes, DLS: Driver-level Separation, \dag used max value, \textcolor{blue}{\( \Delta\)}: gain over the LA-GMF model}
\end{table}

We compared our framework with recent clinical Machine Learning and Deep Learning methods and summarized the results in Table~\ref{tab:clinical_comparison}. Most of these approaches heavily rely on neuroimaging modalities such as MRI and PET, which are costly, invasive, and require significant time and clinical expertise to analyze. In contrast, our scenario-based video framework provides a lightweight, non-invasive alternative and outperforms most clinical ML and DL methods accuracy range of 58–70\%. The most competitive baseline, LA-GMF~\cite{xu2024interpretable}, reported 70.33\% accuracy and 67.22\% F1-score. Our framework reached 72.32\% accuracy and 80.13\% F1-score, showing improvements of $\sim$2\% in accuracy and +12.91\% in F1-score. Additionally, while clinical models often fail to capture how cognitive decline manifests in everyday activities, our method leveraged naturalistic driving behavior to capture functional patterns directly from real-world scenarios. 

Another recent study by Danda et al.~\cite{Danda2025} found that amyloid-positive older adults exhibited higher average speeds, intermittent speed variability and increased acceleration during freeway on-ramp driving (similar to Fwy-int scenario), suggesting less adaptive driving. In clinical research, Amyloid is a protein that accumulates abnormally in the brain and forms plaques linked to the development of MD. Patients with cognitive impairment who test amyloid-positive are typically diagnosed as having MD, whereas amyloid-negative patients are generally excluded from an MD diagnosis~\cite{Souchet2023, bilgel2020amyloid}. These findings confirm that Freeway ramp driving requires increased cognitive load involving behavioral reactions to rapidly changing traffic and abrupt speed shifts, which combine perception with motor control~\cite{Danda2025}. Additionally, they underscore our results of higher classification accuracy (\(\mathbf{a}\) = 71.03\%) and F1-score (\(\mathbf{F1}\) = 0.6712) in Freeway interchange scenarios compared to the Interstate settings (\(\mathbf{a}\) = 55.12\%, \(\mathbf{F1}\) = 0.5368). Although their approach combined driving patterns with physiological signals, it validates the relevance of using contrastive and challenging roadway scenarios in cognitive assessment tasks. Therefore, this study supports our findings that the Freeway interchange scenario provides more informative and strong behavioral cues that the large vision model can leverage to better separate cognitive aging groups.

\subsection{Limitations and Future Works}
The current dataset is constrained to a limited number of driving scenarios, which may not cover the full range of real-world driving complexity (e.g., night driving, inclement weather, heavy traffic). Future work can include a broader set of driving scenarios to generalize findings across real-world traffic and environmental conditions. The driver-level separation was limited to a 3-fold cross-validation with a small pool of test subjects. This can lead to variance in performance estimation and may not generalize well to other populations and road infrastructures. Recruiting more diverse drivers and conducting longitudinal assessments would strengthen the generalizability of the findings. Additionally, the number of dementia subjects in the study was relatively small ($N=9$). Including more dementia participants would guarantee the statistical validity of the findings and provide deeper insights into the driving cues associated with dementia. Moreover, the use of in-vehicle video recording raises potential concerns regarding privacy. Future implementations should address these with robust data management. Another notable limitation is the relatively limited availability of high-quality in-vehicle video data per cognitive group. The model's misclassification rates, particularly in the transitional region, were influenced by the scarcity of this data. We believe that scaling up the dataset with higher-quality video samples for both cognitively normal and dementia groups will allow the method to extract more meaningful and generalizable features. This would be especially crucial for enhancing video analysis accuracy in cases with complex behavioral features.

\section{Conclusion}
\label{sec:Conclusion}
This study highlights the potential of scenario-based driving behavior monitoring as a valuable diagnostic tool for early cognitive decline assessment in older adults.
The results suggest that scenario-based analysis plays a significant role in video-based cognitive assessment. By segmenting driving video data into semantically distinct scenarios, our approach enables targeted analysis of driving behavior under specific cognitive demands. In the scope of this work, the freeway interchange scenario yielded significantly higher classification performance compared to the interstate scenario. Therefore, certain driving scenes inherently capture richer behavioral variance, which large vision models can leverage to distinguish between cognitive aging driving patterns. This approach offers a low-cost, scalable and effective alternative tool to traditional cognitive screening methods, enabling passive and non-intrusive cognitive monitoring. It has direct implications for the healthcare industry, where early detection of cognitive decline remains a costly and time-consuming process. Additionally, the ability to extract digital biomarkers from routine driving behavior transforms personal vehicles into affordable diagnostic tools. This reduces patient burden while supporting continuous and longitudinal cognitive health assessment and monitoring disease progression. Moreover, it holds promise as an efficient screening tool for aging populations, particularly in early detection of cognitive decline, especially in areas with limited access to healthcare facilities.

\section{Acknowledgements}

This work was supported by the National Institutes of Health (NIH), the National Institute on Aging (NIA 5R01AG17177-18), and the University of Nebraska Medical Center Mind \& Brain Health Labs. The views expressed in this paper are those of the authors alone and not the NIH or NIA. We thank our research team for coordinating this project. Due to the presence of personally identifiable information (PII), the data are not available online.

\appendix
\label{app1}

\section{Study Recruitment Criteria}
\label{appendix_a}
The inclusion and exclusion criteria used for participant recruitment in the RWRAD dataset are summarized in Table~\ref{tab:recruitment_criteria}.

\begin{table}[htbp]
\caption{Participant inclusion and exclusion criteria~\cite{chang2025day}}
\label{tab:recruitment_criteria}
\centering
\resizebox{0.90\textwidth}{!}{ 
\small
\begin{tabular}{p{7cm} p{7cm}}
\toprule
\textbf{Inclusion criteria} & \textbf{Exclusion criteria} \\
\midrule
(a) Age between 65 and 90 years & (a) Pulmonary disease requiring chronic medication \\
(b) Residing independently in Nebraska or nearby states (e.g., Iowa) & (b) Congestive heart failure \\
(c) Possession of a valid driver’s license & (c) Major psychiatric illness \\
(d) Meets Nebraska visual acuity standards ($>20/50$ OU, corrected or uncorrected) & (d) Active vestibular disease \\
(e) Proficiency in English & (e) Substance abuse within one year prior to enrollment \\
(f) Capability to provide informed consent & (f) Use of medications that could confound study outcome \\
& (g) Clinical or neurological conditions (e.g., stroke) that could interfere with the study outcome \\
\bottomrule
\end{tabular}
}
\end{table}

\section{Cognitive Tests used for COGSTAT score}
\label{appendix_b}
Cognitive Tests used for calculating COGSTAT score in the RWRAD dataset across five cognitive domains~\cite{jack2018nia} are summarized in Table~\ref{tab:COGSTAT_tests}.

\begin{table}[h]
\centering
\caption{Cognitive tests for the COGSTAT composite score.}
\label{tab:COGSTAT_tests}
\resizebox{0.90\textwidth}{!}{ 
\small
\begin{tabular}{p{5cm} p{8cm}}
\toprule
\textbf{Cognitive Domain} & \textbf{Cognitive Test} \\
\midrule
(1) Memory & Benson, Recall \\
       & Craft Story Delay, Verbatim \\
       & HVLT, Delay \\
\midrule
(2) Language & Category Fluency, Animals \\
         & Category Fluency, Vegetables \\
         & MINT  \\
\midrule
(3) Visuospatial & Benson, Copy  \\
             & WAIS Blocks\\
\midrule
(4) Executive Function & Trails B\\
                   & Verbal Fluency (F) \\
                   & Verbal Fluency (L) \\
\midrule
(5) Attention & Trails A \\
          & Number Span Forward (Total correct)\\
          & Number Span Backward (Total correct) \\
\bottomrule
\end{tabular}
}
\caption*{\small Note: If all test scores $=$ mean score, then the final score will be $14\times 50 = 700$}
\end{table}

\bibliographystyle{elsarticle-num} 
\bibliography{main}






\end{document}